\documentclass[runningheads,a4paper]{llncs}

\usepackage{amssymb}
\usepackage{amsmath}
\usepackage{graphicx}
\global\hyphenpenalty=0

\usepackage{subcaption}

\usepackage{url}
\urldef{\mailsa}\path|{nccodell
}@us.ibm.com|    
\urldef{\mailsb}\path|{rotembev,marchetm,duszas,kallooa,liopyrik,halperna
}@mskcc.org| 
\urldef{\mailsc}\path|{dgutman
}@emory.edu| 
\urldef{\mailsd}\path|{ecelebi
}@uca.edu| 
\urldef{\mailse}\path|{brian.helba
}@kitware.com| 
\urldef{\mailsf}\path|{philipp.tschandl,harald.kittler
}@meduniwein.ac.at| 
\newcommand{\keywords}[1]{\par\addvspace\baselineskip
\noindent\keywordname\enspace\ignorespaces#1}

\begin{document}

\mainmatter  

\title{Skin Lesion Analysis Toward Melanoma Detection 2018: A Challenge Hosted by the International Skin Imaging Collaboration (ISIC) }

\titlerunning{Skin Lesion Analysis Toward Melanoma Detection 2018}

%
%

\author{Noel Codella$^1$ \and Veronica Rotemberg$^2$ \and Philipp Tschandl$^3$ \and M. Emre Celebi$^4$ \and Stephen Dusza$^2$ \and David Gutman$^5$  \and Brian Helba$^6$ \and Aadi Kalloo$^2$ \and Konstantinos Liopyris$^2$ \and Michael Marchetti$^2$ \and Harald Kittler$^3$$^*$ \and Allan Halpern$^2$$^*$}

\authorrunning{Codella et al.  }

\institute{ $^1$IBM T.J. Watson Research Center, Yorktown Heights, NY, USA \\
$^2$Memorial Sloan-Kettering Cancer Center, New York, NY, USA \\
$^3$Medical University of Vienna, Vienna, Austria \\
$^4$University of Central Arkansas, Conway, AR, USA \\
$^5$Emory University, Atlanta, GA, USA \\
$^6$Kitware, Inc., Clifton Park, NY, USA \\ 
$^*$Co-Investigators
}

\toctitle{Skin Lesion Analysis Toward Melanoma Detection 2018: A Challenge Hosted by the International Skin Imaging Collaboration (ISIC)}
\tocauthor{Codella et al.}
\maketitle

\begin{abstract}
This work summarizes the results of the largest skin image analysis challenge in the world, hosted by the International Skin Imaging Collaboration (ISIC), a global partnership that has organized the world's largest public repository of dermoscopic images of skin. The challenge was hosted in 2018 at the Medical Image Computing and Computer Assisted Intervention (MICCAI) conference in Granada, Spain. The dataset included over 12,500 images across 3 tasks. 900 users registered for data download, 115 submitted to the lesion segmentation task, 25 submitted to the lesion attribute detection task, and 159 submitted to the disease classification task. Novel evaluation protocols were established, including a new test for segmentation algorithm performance, and a test for algorithm ability to generalize. Results show that top segmentation algorithms still fail on over 10\% of images on average, and algorithms with equal performance on test data can have different abilities to generalize. This is an important consideration for agencies regulating the growing set of machine learning tools in the healthcare domain, and sets a new standard for future public challenges in healthcare.

\keywords{deep learning, dermoscopy, melanoma, skin cancer}
\end{abstract}

\section{Introduction}

Skin cancer is the most common form of cancer in the United States, with the annual cost of care exceeding \$8 billion \cite{stat2}. With early detection, the 5 year survival rate of the most deadly form, melanoma, can be up to 99\%; however, delayed diagnosis causes the survival rate to dramatically decrease to 23\% \cite{stat3}. 

Due to the importance of early detection, much work has been dedicated to increasing the accuracy and scale of diagnostic methods. In 2016 and 2017, the International Skin Imaging Collaboration (ISIC), a global partnership that has organized the world's largest repository of publicly available dermoscopic images, hosted the first public benchmarks for melanoma detection in dermoscopic images, titled ``Skin Lesion Analysis Towards Melanoma Detection'', at the IEEE International Symposium of Biomedical Imaging (ISBI) \cite{isic2016,jaadarticle,isbi2017}. The two consecutive challenges attracted global participation, with over 900 registrations and over 350 submissions, making them the largest standardized and comparative studies at the time, yielding novel findings and numerous publications, and have been tacitly accepted as a de-facto reference standard by other groups \cite{codellajrd,nature,recod,montypython}.

This article describes the methods and the results from the most recent instance of the ISIC Challenge on Skin Lesion Analysis Towards Melanoma Detection, hosted in 2018 at the Medical Image Computing and Computer Aided Intervention (MICCAI) conference in Granada, Spain. In addition to considerable increases in the size of the dataset and number of diagnostic labels, key changes to evaluation criteria and study design were implemented to better reflect the complexity of clinical scenarios encountered in practice. These changes included 1) a new segmentation metric to better account for extreme deviations from interobserver variability, 2) implementation of balanced accuracy for classification decisions to minimize influence of prevalence and prior distributions that may not be consistent in practice, and 3) inclusion of external test data from institutions excluded from representation in the training dataset, to better assess how algorithms generalize beyond the environments for which they were trained. The impacts of each change are examined in this work, followed by a set of recommendations for future challenges.

\section{Methods}
\label{methods}

\begin{figure}[t!]
  \centerline{\includegraphics[width=12cm]{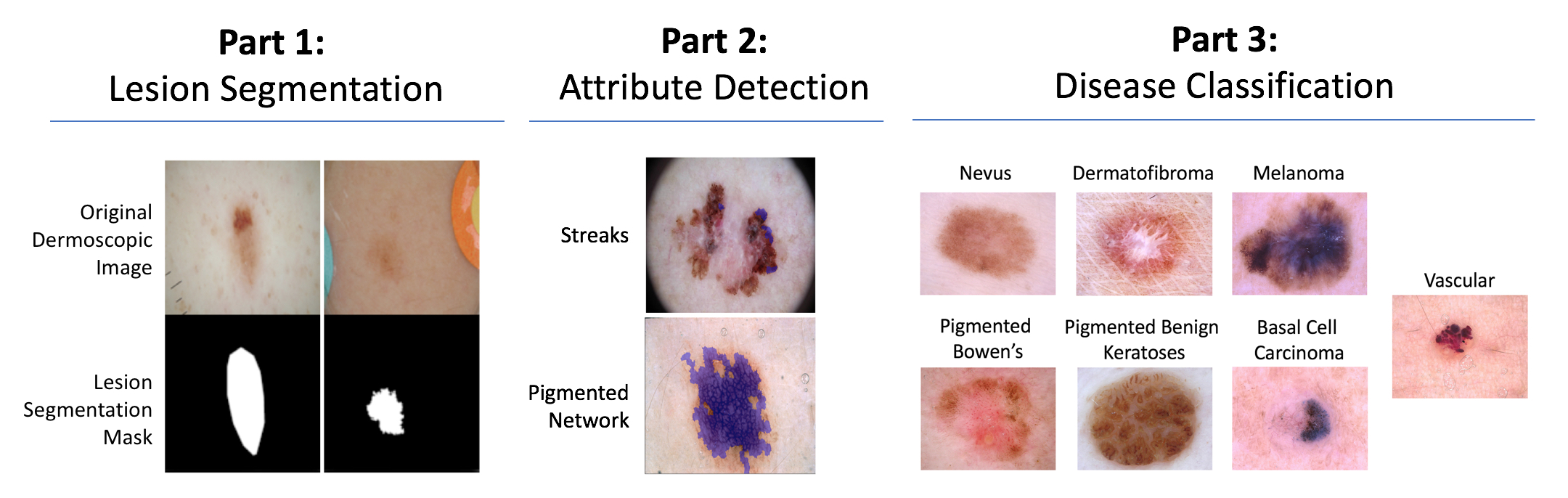}}
  \caption{ Example training data and ground truth from Part 1: Lesion Segmentation, Part 2: Attribution Detection, and Part 3: Disease Classification.  }
\label{fig:examples}
\end{figure}

The challenge was separated into 3 image analysis tasks of lesion segmentation, attribute detection, and disease classification (Fig. \ref{fig:examples}). There was no requirement that any task be independent of the data or analytics developed for the other tasks. Participants were required to provide 4-page manuscripts along with submissions describing implemented methods. Participants were allowed to use alternative sources of in-domain (dermoscopic) data, but were required to disclose such use in a specific meta-data field. Use of out-of-domain data (non-dermoscopic), such as ImageNet, was expected to be mentioned in manuscripts, but not required to be disclosed in a separate meta-data field.

\subsection{Part 1: Lesion Segmentation}

For Part 1, 2,594 dermoscopic images with ground truth segmentation masks were provided for training. For validation and test sets, 100 and 1,000 images were provided, respectively, without ground truth masks.

Evaluation criteria historically has been the Jaccard index ~\cite{isic2016,jaadarticle,isbi2017}, averaged over all images in the dataset. In practice, ground truth segmentation masks are influenced by inter-observer and intra-observer variability, due to variations in human annotators and variations in annotation software \cite{segpaper}. An ideal evaluation would generate several ground truth segmentation masks for every image using multiple annotators and software systems. Then, for each image, predicted masks would be compared to the multiple ground truth masks to determine whether the predicted mask falls outside or within observer variability. However, this would multiply the manual labor required to generate ground truth masks, rendering such an evaluation impractical and infeasible.

As an approximation to this ideal evaluation criteria, we introduced ``Thresholded Jaccard'', which works similarly to standard Jaccard, with one important exception: if the Jaccard value of a particular mask falls below a threshold {\em T}, the Jaccard is set to zero. The value of the threshold {\em T} defines the point in which a segmentation is considered ``incorrect''.

Prior work measured the average Jaccard between 3 expert annotators on 100 images from the 2016 challenge \cite{codellajrd}. The resulting values were 0.743, 0.754, and 0.861, yielding an average of 0.786 and a range of 0.118. For the 2018 challenge, {\em T} was defined as 0.65, by rounding the lowest agreement to 0.75 and subtracting one rounded range (0.10) to increase the certainty (specificity) of failure.

\subsection{Part 2: Lesion Attribute Detection}

For Part 2, 2,594 images with 12,970 ground truth segmentation masks for 5 attributes were provided for training. For validation and held-out test sets, 100 and 1,000 images were provided, respectively, without masks.

Jaccard was used as the evaluation metric this year in order to facilitate possible re-use of methods developed for segmentation, and encourage greater participation. As some dermoscopic attributes may be entirely absent from certain images, the Jaccard value for such attributes is ill-defined (division by 0). To overcome this difficulty, the Jaccard was measured by computing the TP, FP, and FN for the entire dataset, rather than a single image at a time.

\subsection{Part 3: Lesion Disease Classification}

For Part 3, 10,015 dermoscopic images with 7 ground truth classification labels were provided for training \cite{ham10k}. For validation and held-out test sets, 193 and 1,512 images were provided, respectively, without ground truth.

Held-out test data was further split into two partitions: 1) an {\em ``internal''} partition, consisting of 1,196 images selected from data sources that were consistent with the training dataset (two institutions in Austria and Australia), and 2) an {\em ``external''} partition, consisting of 316 images additionally selected from data sources not reflected in the training dataset (institutions from Turkey, New Zealand, Sweden, and Argentina). 

Evaluation was carried out using balanced accuracy (mean recall across classes after mutually exclusive classification decision), because dataset prevalence may not be reflective of real world disease prevalence, especially with regard to over-representation of melanomas. Previous years had used melanoma average precision \cite{isic2016} and melanoma AUC (area under receiver operating characteristic curve), which are only robust to the prevalence imbalance of melanomas, and may be influenced by clinically irrelevant low-sensitivity performance.

\section{Results}
\label{results}

\subsection{Part 1: Lesion Segmentation}

\begin{figure}[t!]
  \centerline{\includegraphics[width=12cm]{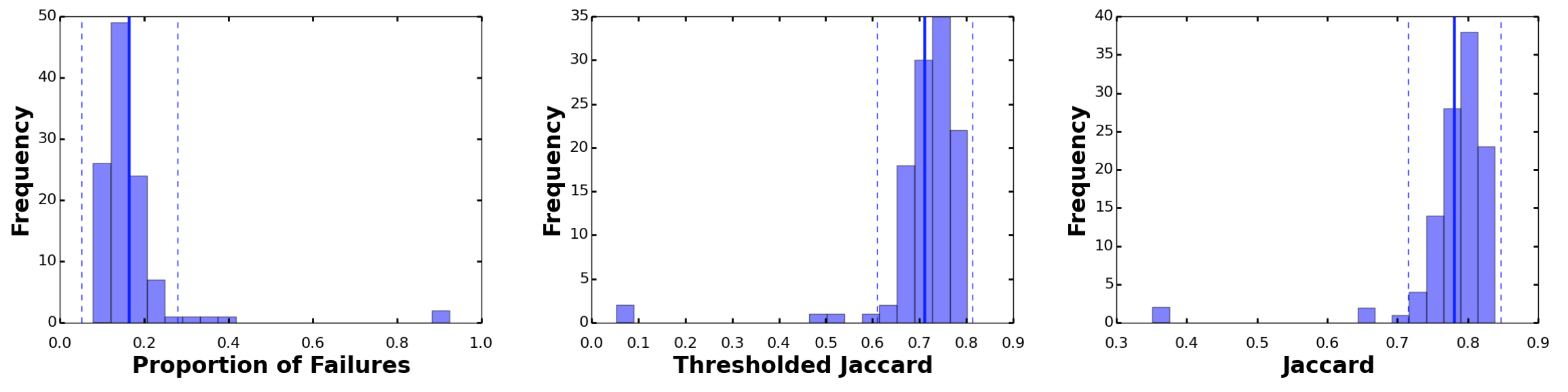}}
  \caption{ Histograms of submissions for Part 1: Lesion Segmentation. Performance on the X-axis, and number of submissions on the Y-axis. Average values shown as solid vertical lines, and +/- standard deviation shown as dotted vertical lines.}
\label{fig:seg_hist}
\end{figure}

\begin{figure}[h!]
  \centerline{\includegraphics[width=12cm]{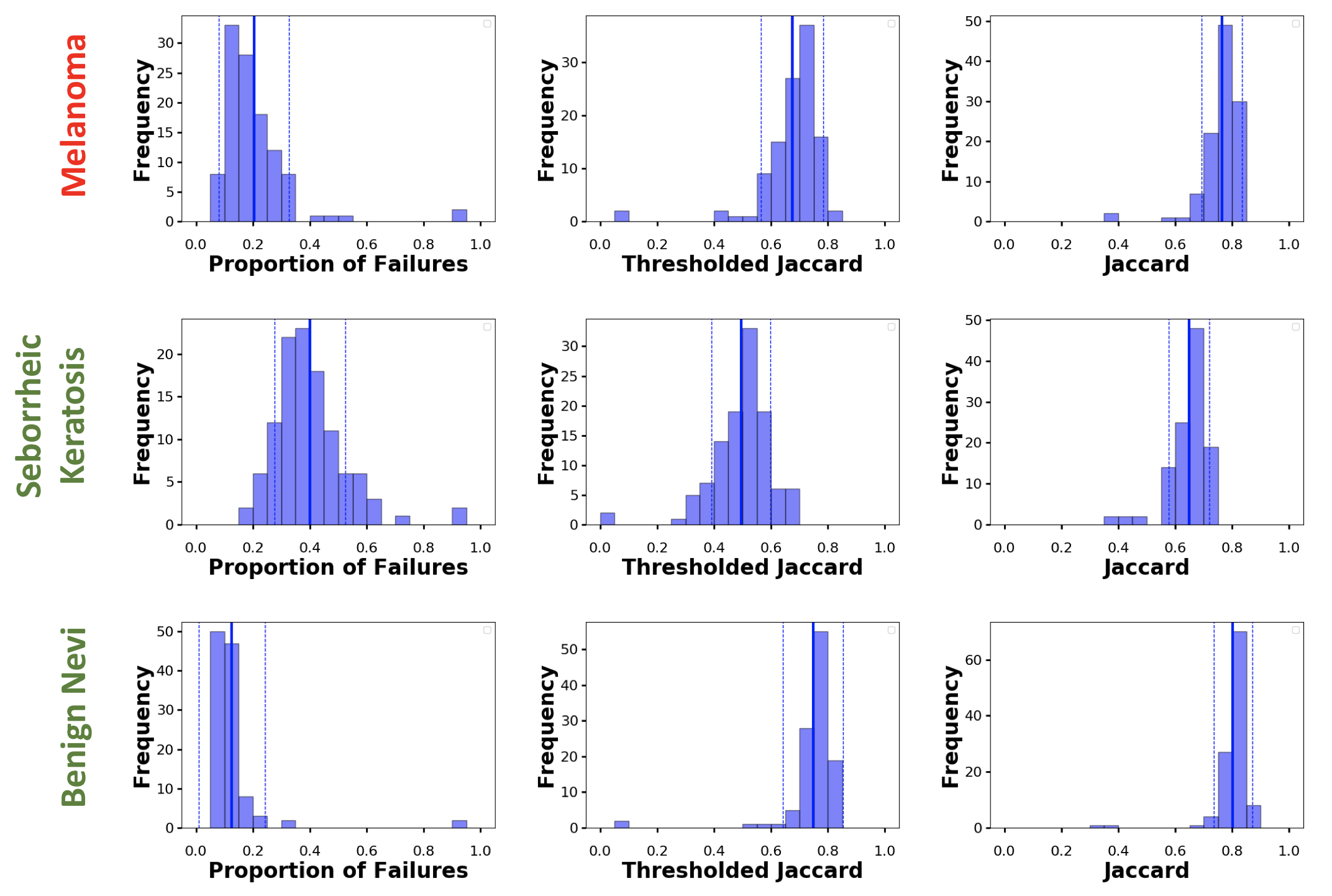}}
  \caption{Histograms of submissions for Part 1: Lesion Segmentation, stratified by disease state. Performance on the X-axis, and number of submissions on the Y-axis. Average values shown as solid vertical lines, and +/- standard deviation shown as dotted vertical lines.}
\label{fig:seg_hist_disease}
\end{figure}

In total, 112 submissions were received for Part 1. The top performing submission achieved a Thresholded Jaccard of 0.802, with many of the other top algorithms also achieving approximately 0.8.  A histogram summary of the submissions are shown in Fig. \ref{fig:seg_hist}, showing the proportion of failures (segmentations below 0.65 Jaccard), performance according to Thresholded Jaccard, and performance according to Jaccard. Supplemental Fig. \ref{fig:seg_hist_disease} shows the histograms stratified by disease type, and supplemental Table \ref{segvals} shows details of the top 5 submissions. What is clear from this analysis is that even though submissions may achieve very high scores in average Jaccard values (over 0.8, exceeding even previous reports of average Jaccard values for inter-observer variability of 0.786 \cite{codellajrd}), most methods still fail to properly segment over 10\% of images (failures most common for seborrheic keratoses). This is an important observation that is often diluted by most aggregated statistics.

Fig. \ref{fig:seg_rank}A shows the correlation between both Thresholded Jaccard and Jaccard against the proportion of segmentation failures, demonstrating that Thresholded Jaccard has a correlation slope closer to 1,  suggesting that the new metric may be a better assessment of clinical utility. Fig. \ref{fig:seg_rank}B shows a scatter plot of participant challenge rank according to Thresholded Jaccard (X-axis) and Jaccard (Y-axis), demonstrating that changing the evaluation criteria to Threhsolded Jaccard has an impact on the ranking of participant algorithms. 

\begin{figure*}[t!]
\begin{subfigure}{.5\textwidth}
  \caption{A)}
  \includegraphics[width=6cm]{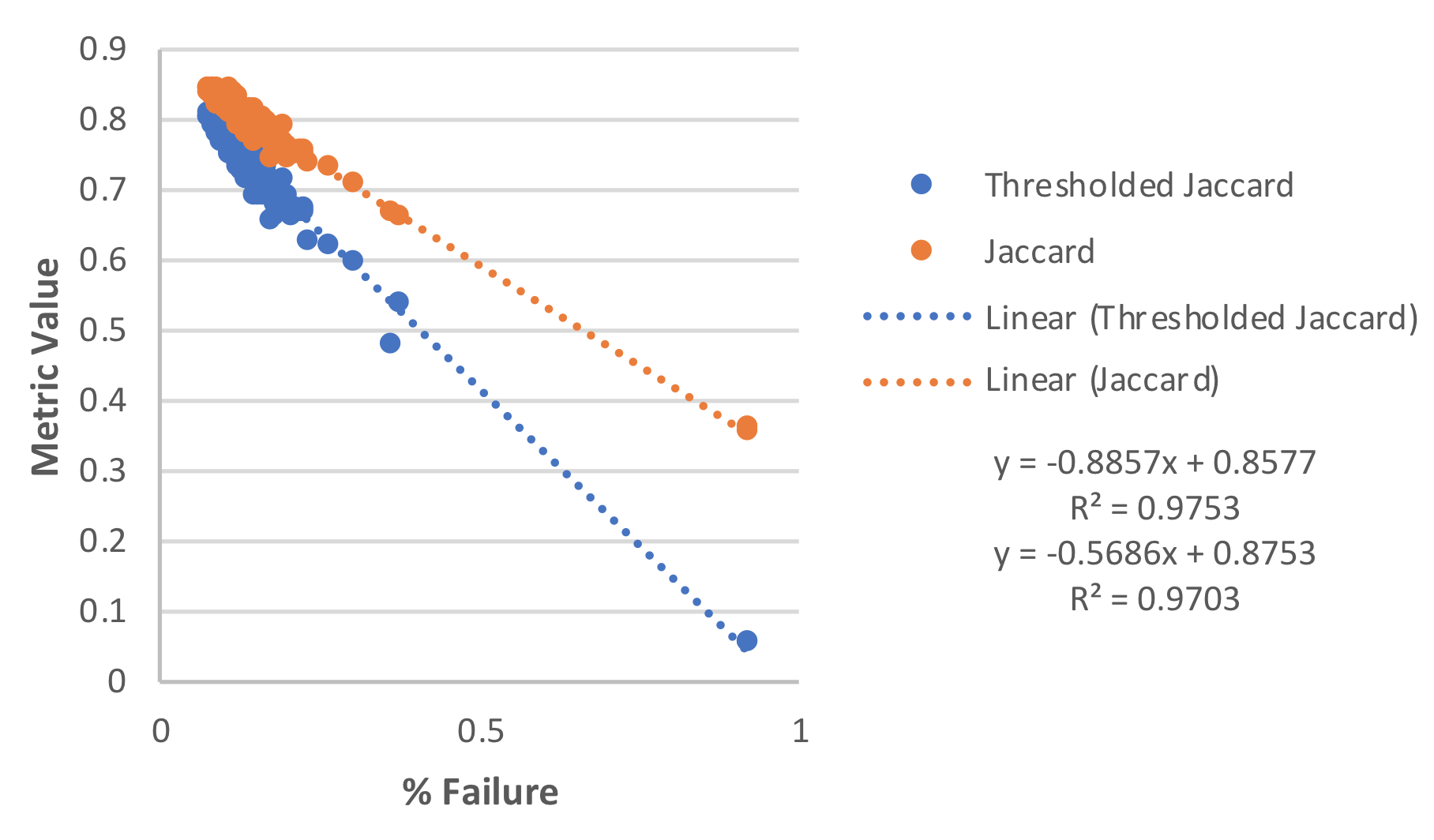}
  \label{fig:sfig1}
\end{subfigure}%
\begin{subfigure}{.5\textwidth}
  \caption{B)}
  \includegraphics[width=6cm]{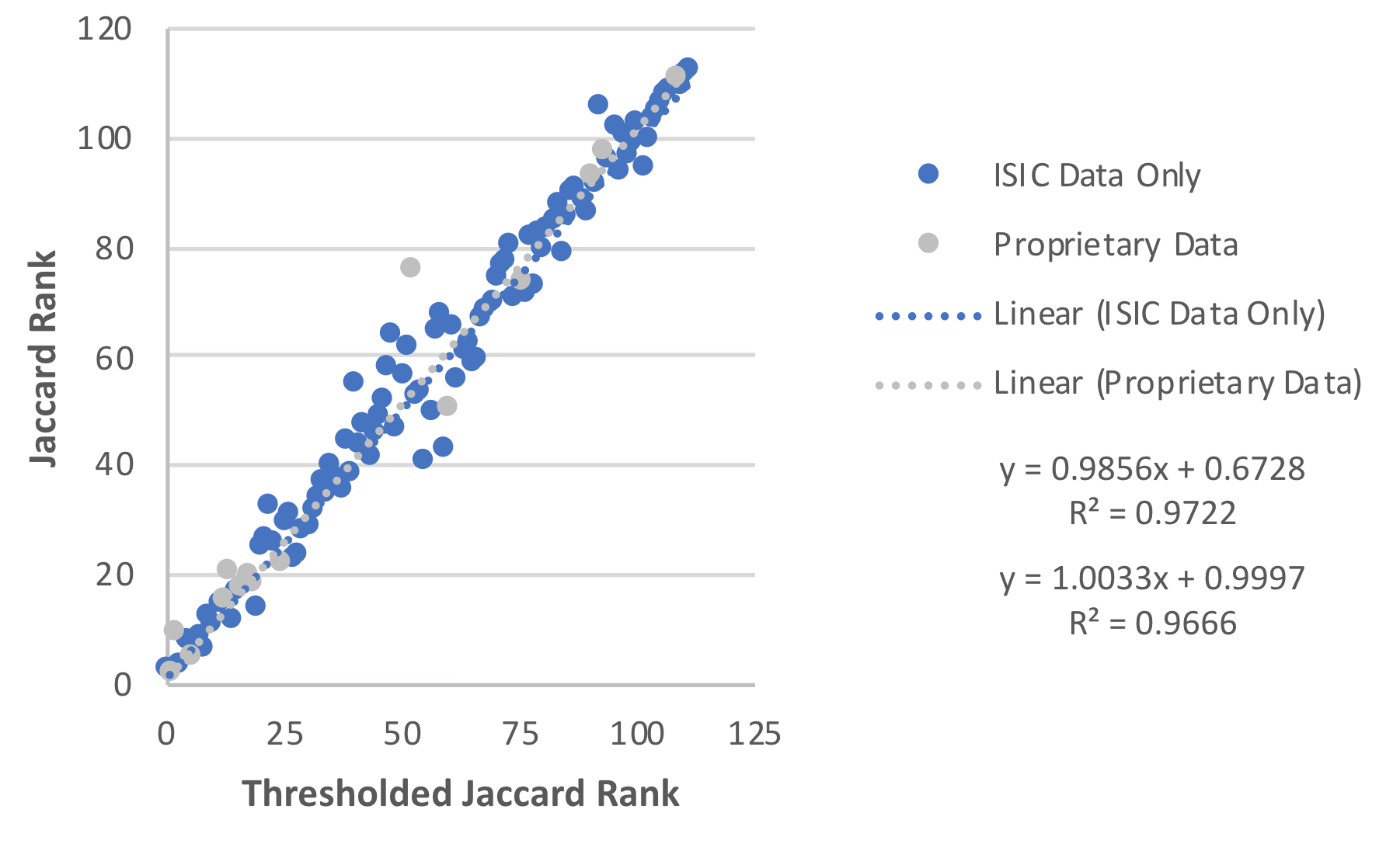}
  \label{fig:sfig1}
\end{subfigure}%
\addtocounter{figure}{-1}
\caption{Assessment of new Threshold Jaccard metric. A) Proportion of segmentation failures (X-axis) vs. various metric values (Y-axis). B) Participant ranking by Thresholded Jaccard (X-axis) vs. Jaccard (Y-axis).  }
\label{fig:seg_rank}
\end{figure*}

\subsection{Part 2: Lesion Attribute Detection}

\begin{figure}[t!]
  \centerline{\includegraphics[width=12.5cm]{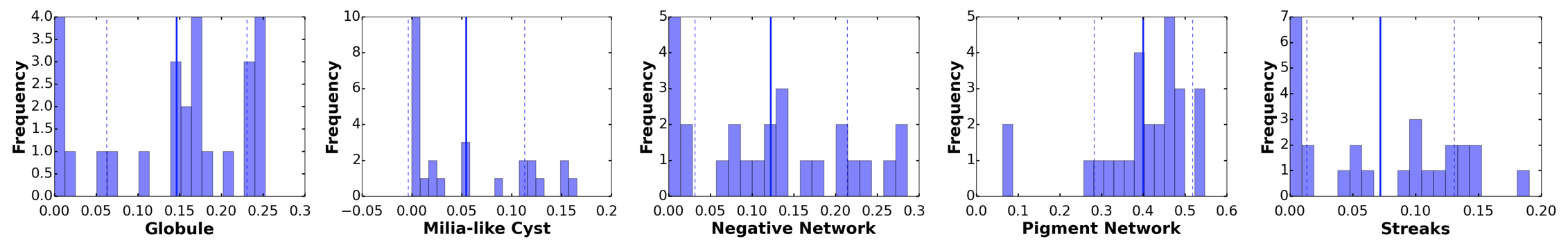}}
  \caption{ Histogram of submissions to Part 2: Lesion Attribute Detection.}
\label{fig:attr}
\end{figure}

In total, 26 submissions were received for Part 2. The histograms of performance of each attribute is plotted in Fig. \ref{fig:attr} according to average Jaccard. The distribution of values were exceptionally low, with the best submissions achieving an average of 0.473. Poor performance may be the result of several factors, which may include that dermoscopic attributes tend to have poor inter-observer correlation among expert clinicians \cite{correlation}. The implications for future challenges may be that either the field of clinical dermoscopic attributes must mature further before additional research is performed to apply machine learning---or that machine learning methods should be applied to the reverse problem: to help find and annotate specific patterns that may correlate strongly with disease. 

\subsection{Part 3: Lesion Disease Classification}

\begin{figure}[t!]
  \centerline{\includegraphics[width=12cm]{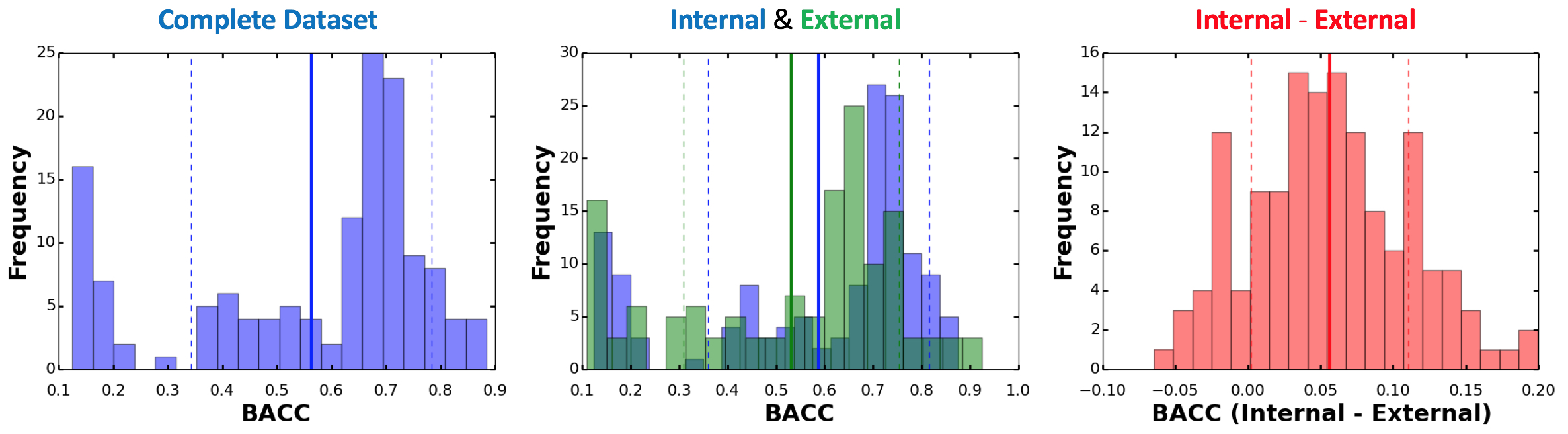}}
  \caption{ Histograms of submissions for Part 3: Lesion Classification. Average values showed as solid vertical lines, and +/- standard deviation showed as dotted vertical lines. {\em Left:} Entire test dataset. {\em Center:} Internal (blue) and External (green) test partitions split. {\em Right:} Histogram of the subtraction of external test performance from internal. }
\label{fig:class_hist}
\end{figure}

In total, 141 submissions were received to Part 3. Histograms of submission performance, according to balanced accuracy (BACC), are shown in Fig. \ref{fig:class_hist}. ROC plots for the top 12 submissions are shown in Supplemental Fig. \ref{fig:roc_disease}, details of top 5 submissions are shown in Supplemental Table \ref{diagvals}, and histograms for AUC by disease are shown in Supplemental Fig. \ref{fig:class_hist}. The highest performance achieved was 0.885. Correlation between internal and external test dataset performance for each submission is shown in Fig. \ref{fig:class_source}A, and between whole test set performance and the difference between internal and external sets in \ref{fig:class_source}B. Differences in ranking according to balanced accuracy, accuracy, and mean AUC are plotted in Fig. \ref{fig:class_rank}. 

These analyses provide the following important observations: 1) Most submissions overfit and perform better on internal data vs. external data (Fig. \ref{fig:class_hist}), but some approaches, including the top performers, do not (Figs. \ref{fig:class_hist} \& \ref{fig:class_source}B). 2) The use of proprietary data is not required in order to prevent overfitting to internal data (Fig. \ref{fig:class_source}B). 3) Various algorithms may achieve similar whole test set performance but vary widely on their ability to generalize (Fig. \ref{fig:class_source}B). 4) Simple linear correlation between internal and external test dataset performance does not elucidate the spread of overfitting as clearly as plotting by the difference between datasets (Fig. \ref{fig:class_source}A vs. Fig. \ref{fig:class_source}B). 5) The choice of evaluation metric has a significant impact on participant ranking (Fig. \ref{fig:class_rank}). Use of balanced accuracy is critical to select the best unbiased classifier, rather than one that overfits to arbitrary dataset prevalence, as is the case with accuracy (Fig. \ref{fig:class_rank}A). Even other unbiased estimators, such as mean AUC (Fig. \ref{fig:class_rank}B) show significant differences in rank as compared to balanced accuracy, and consider areas of operating curves, such as low-recall regions, that may not be clinically relevant.   

\begin{figure}[t!]
\begin{subfigure}{.5\textwidth}
  \centering
  \caption{A)}
  \includegraphics[width=6cm]{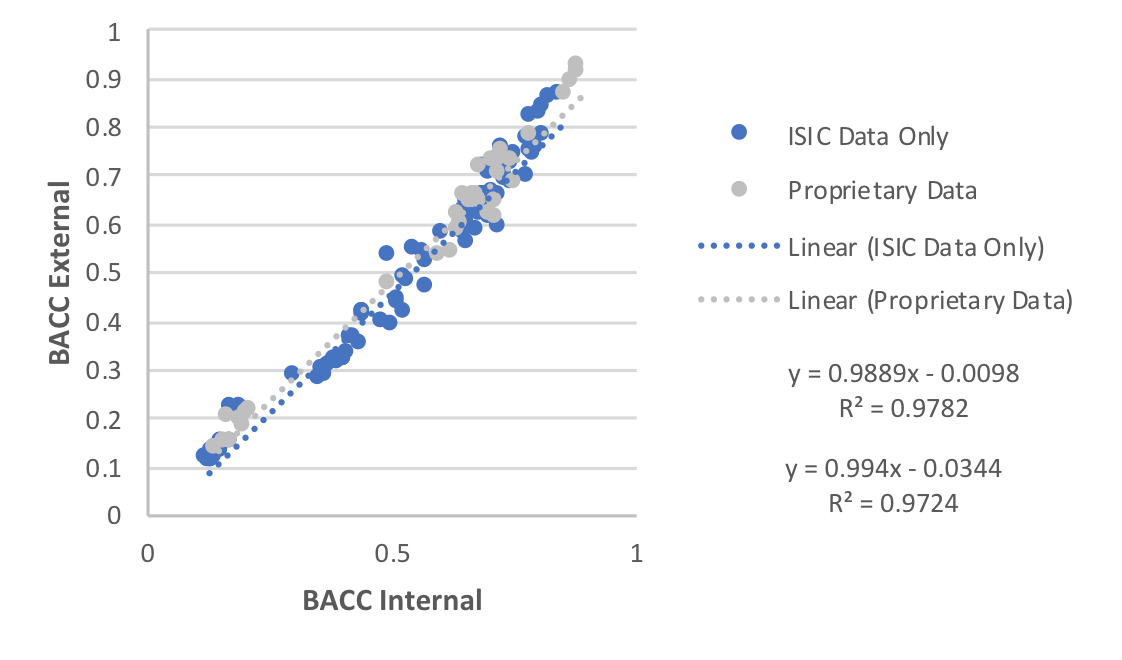}
  
  \label{fig:sfig1}
\end{subfigure}%
\begin{subfigure}{.5\textwidth}
  \centering
  \caption{B)}
  \includegraphics[width=6cm]{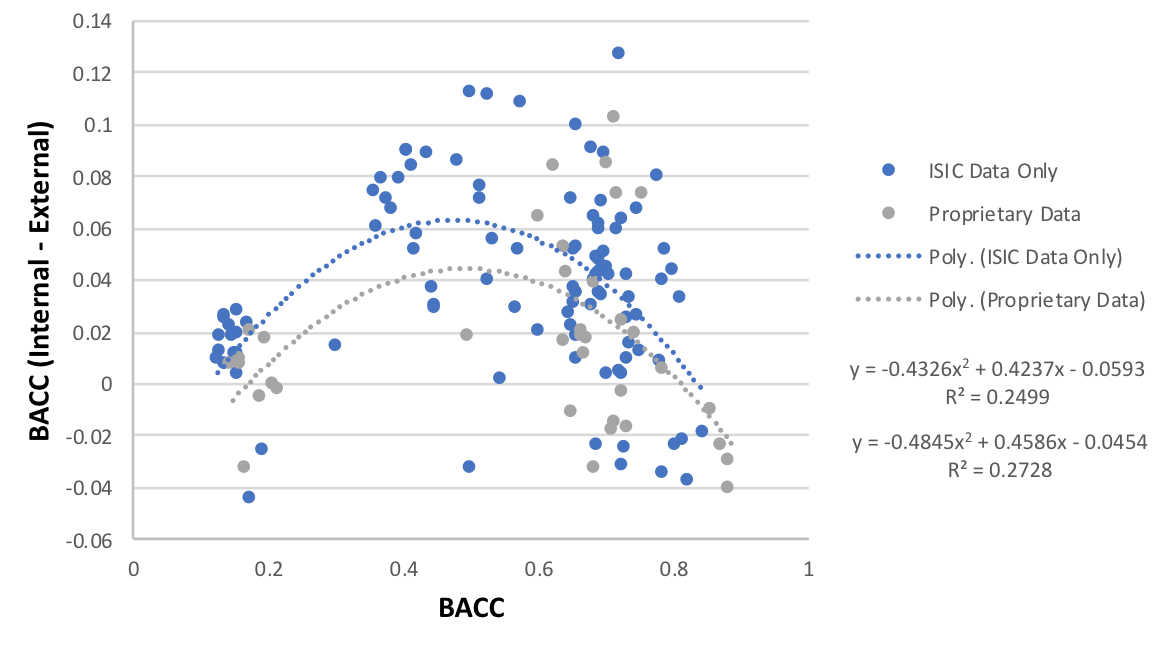}
  
  \label{fig:sfig2}
\end{subfigure}
\addtocounter{figure}{-1}
\caption{Comparisons of participant performance on internal test dataset vs. external test dataset. {\em A:} Internal vs. external test set peformance. {\em B:} Whole test set vs. the difference between internal and external test set performances.  }
\label{fig:class_source}
\end{figure}

\begin{figure}[t!]
\begin{subfigure}{.5\textwidth}
  \centering
  \caption{A)}
  \includegraphics[width=6cm]{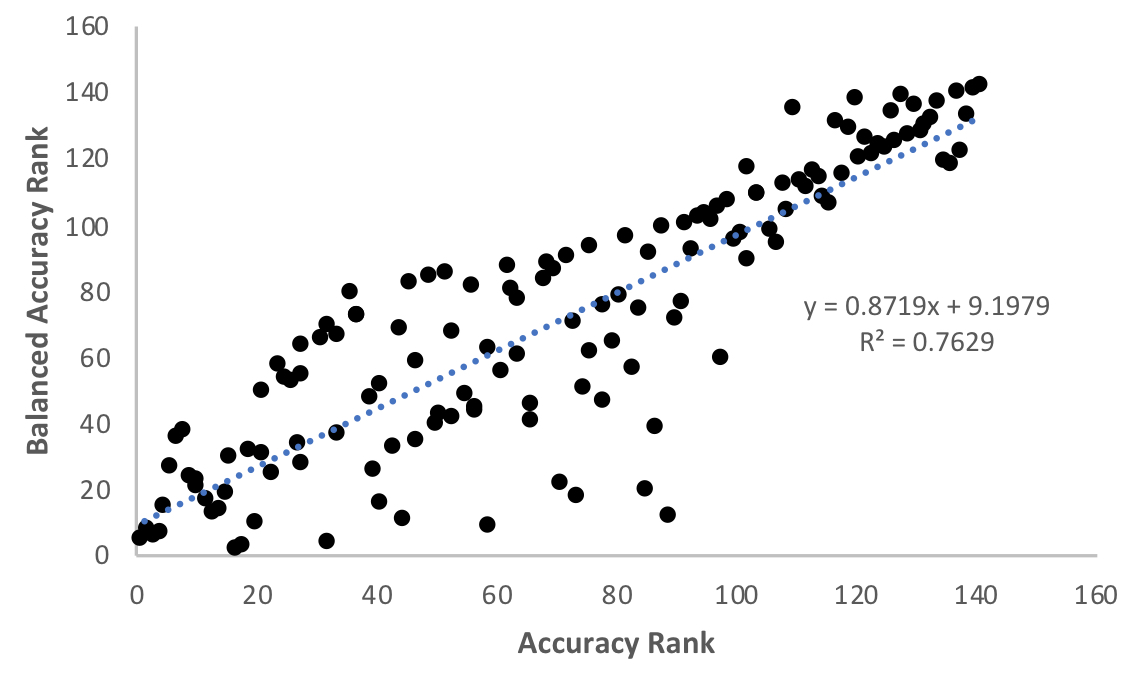}
  
  \label{fig:sfig1}
\end{subfigure}%
\begin{subfigure}{.5\textwidth}
  \centering
  \caption{B)}
  \includegraphics[width=6cm]{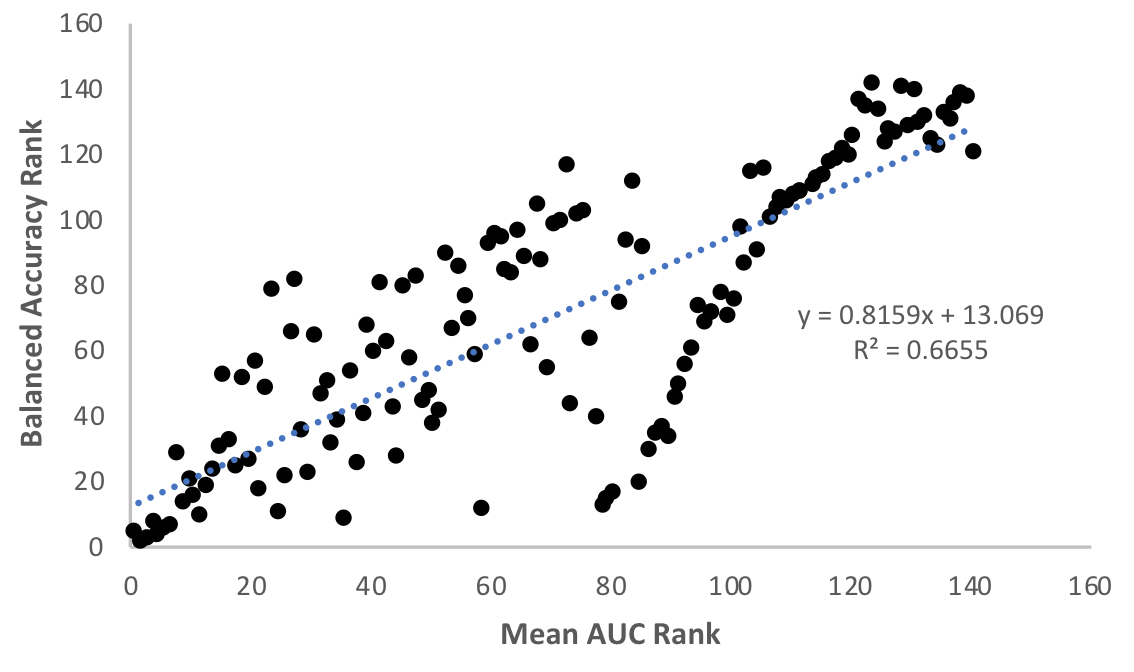}
  
  \label{fig:sfig2}
\end{subfigure}
\addtocounter{figure}{-1}
\caption{Impact of new balanced accuracy (BACC) jaccard metric on participant ranking. A) BACC vs. ACC. B) BACC vs. Mean AUC }
\label{fig:class_rank}
\end{figure}

\section{Discussion \& Conclusion}
\label{conclusion}

This work summarizes the results of the 2018 MICCAI Challenge on Skin Lesion Analysis Toward Melanoma Detection, hosted by the International Skin Imaging Collaboration (ISIC), which represents the de facto standard benchmark for machine learning in this domain. In total, over 12,500 images were made available for training across 3 tasks, and over 2,000 images for testing. 900 teams registered, and 299 submissions were received, making this challenge the largest in the field to date, in terms of size, complexity, and degree of participation. 

Several changes in evaluation criteria were implemented in comparison to previous ISIC challenges to better reflect difficulties encountered in clinical practice. These include 1) Thresholded Jaccard, which severely penalizes segmentations that fall outside an estimate of interobserver variability, 2) balanced accuracy, which avoids encouraging classification systems from over-fitting potential dataset imbalances, and 3) a dual-partition held-out test set, including data sourced from institutions not reflected in the training dataset, to better measure algorithm ability to generalize. 

Results show that 1) Thresholded Jaccard better captures the proportion of segmentation failures in comparison to Jaccard, 2) balanced accuracy leads to significant changes in participant ranking versus other metrics that may be more prone to imbalance or influence from clinically irrelevant ROC regions, 3) multi-partition test sets containing data not reflected in training dataset are an effective way to differentiate the ability of algorithms to generalize, and 4) poor performance observed in Part 2 may imply that dermoscopic attributes mature further before research is continued to apply machine learning. 

Future challenges and regulatory agencies in medical imaging and dermoscopic image analysis should consider the presented evaluation criteria to best quantify algorithm performance, robustness, and ability to generalize in clinical scenarios. 


\pagebreak

\newpage
\section{Supplementary Material}
\label{supp}

\begin{table}[]
\centering
\setlength{\tabcolsep}{3pt}
\renewcommand{\arraystretch}{1.1}
\begin{tabular}{|c|c|c|c|c|c|c|c|c|c|c|c|c|}
\hline
& \multicolumn{3}{|c|}{\bf ALL} & \multicolumn{3}{|c|}{\bf MEL} & \multicolumn{3}{|c|}{\bf SEBK} & \multicolumn{3}{|c|}{\bf NEVI} \\
\hline 
\bf R & \bf F & \bf TJ & \bf J & \bf F & \bf TJ & \bf J & \bf F & \bf TJ & \bf J & \bf F & \bf TJ & \bf J \\
\hline
1 & 0.093 & 0.802 & 0.838 & 0.095 & 0.792 & 0.832 & 0.310 & 0.577 & 0.698 & 0.066 & 0.832 & 0.856 \\
2 & 0.079 & 0.801 & 0.838 & 0.090 & 0.782 & 0.830 & 0.195 & 0.667 & 0.743 & 0.063 & 0.820 & 0.851 \\
3 & 0.083 & 0.799 & 0.834 & 0.100 & 0.782 & 0.826 & 0.299 & 0.585 & 0.706 & 0.053 & 0.829 & 0.852 \\
4 & 0.085 & 0.798 & 0.838 & 0.090 & 0.792 & 0.839 & 0.207 & 0.656 & 0.740 & 0.069 & 0.817 & 0.848 \\
5 & 0.084 & 0.796 & 0.837 & 0.095 & 0.799 & 0.849 & 0.195 & 0.670 & 0.738 & 0.067 & 0.811 & 0.845 \\
\hline
\end{tabular}
\caption{\label{segvals}Top 5 submissions to Part 1: Lesion Segmentation. ALL = Performance on entire test set. MEL = Performance on Melanomas. SEBK = Performance on Seborrheic Keratoses. NEVI = Performance on Benign Nevi. R = Rank. F = Failure rate. TJ = Thresholded Jaccard. J = Jaccard. }
\end{table}

\begin{table}[]
\centering
\setlength{\tabcolsep}{3pt}
\renewcommand{\arraystretch}{1.1}
\begin{tabular}{|c|c|c|c|c|c|c|c|c|c|c|}
\hline
& & \multicolumn{2}{|c|}{\bf Accuracy} & \multicolumn{7}{|c|}{\bf AUC} \\
\hline
& \bf R & \bf ACC & \bf BACC & \bf MEL & \bf NV & \bf BCC & \bf AKIEC & \bf BKL & \bf DF & \bf VASC \\
\hline
& \bf 1 & 0.851 & 0.885 & 0.949 & 0.979 & 0.997 & 0.987 & 0.974 & 0.992 & 1.000 \\
& \bf 2 & 0.850 & 0.882 & 0.946 & 0.981 & 0.997 & 0.985 & 0.977 & 0.990 & 1.000 \\
\bf ALL & \bf 3 & 0.827 & 0.871 & 0.948 & 0.978 & 0.996 & 0.981 & 0.971 & 0.986 & 0.999 \\
& \bf 4 & 0.896 & 0.856 & 0.959 & 0.983 & 0.995 & 0.995 & 0.990 & 0.987 & 0.999 \\
& \bf 5 & 0.884 & 0.845 & 0.945 & 0.974 & 0.992 & 0.988 & 0.969 & 0.982 & 0.998 \\
\hline 
& \bf 1 & 0.841 & 0.875 & 0.945 & 0.978 & 0.997 & 0.986 & 0.969 & 0.992 & 1.000 \\
& \bf 2 & 0.842 & 0.875 & 0.941 & 0.980 & 0.997 & 0.983 & 0.972 & 0.993 & 0.999 \\
\bf INT & \bf 3 & 0.820 & 0.875 & 0.944 & 0.977 & 0.997 & 0.980 & 0.965 & 0.994 & 0.999 \\
& \bf 4 & 0.907 & 0.854 & 0.961 & 0.982 & 0.999 & 0.995 & 0.988 & 0.973 & 0.999 \\
& \bf 5 & 0.894 & 0.841 & 0.939 & 0.973 & 0.997 & 0.988 & 0.963 & 0.973 & 0.998 \\
\hline 
& \bf 1 & 0.886 & 0.925 & 0.970 & 0.984 & 0.998 & 1.000 & 0.984 & 0.993 & 1.000 \\
& \bf 2 & 0.880 & 0.911 & 0.966 & 0.986 & 0.998 & 1.000 & 0.987 & 0.989 & 1.000 \\
\bf EXT & \bf 3 & 0.854 & 0.894 & 0.970 & 0.982 & 0.993 & 1.000 & 0.983 & 0.984 & 0.999 \\
& \bf 4 & 0.854 & 0.866 & 0.984 & 0.984 & 0.987 & 0.997 & 0.994 & 0.995 & 1.000 \\
& \bf 5 & 0.848 & 0.864 & 0.976 & 0.971 & 0.978 & 0.998 & 0.978 & 0.984 & 0.999 \\
\hline
\end{tabular}
\caption{\label{diagvals}Top 5 submissions to Part 3: Lesion Classification. ALL = Performance on entire test set. INT = Performance on Internal Test Set. EXT = Performance on External Test Set. R = Rank. ACC = Accuracy. BACC = Balanced Accuracy. MEL = Melanoma. NV = Nevi. BCC = Basal cell carcinoma. AKIEC = Actinic keratosis. BKL = Benign keratosis. DF = Dermatofibroma. VASC = Vascular Lesion. }
\end{table}

\begin{figure}[h!]
  \centerline{\includegraphics[width=12cm]{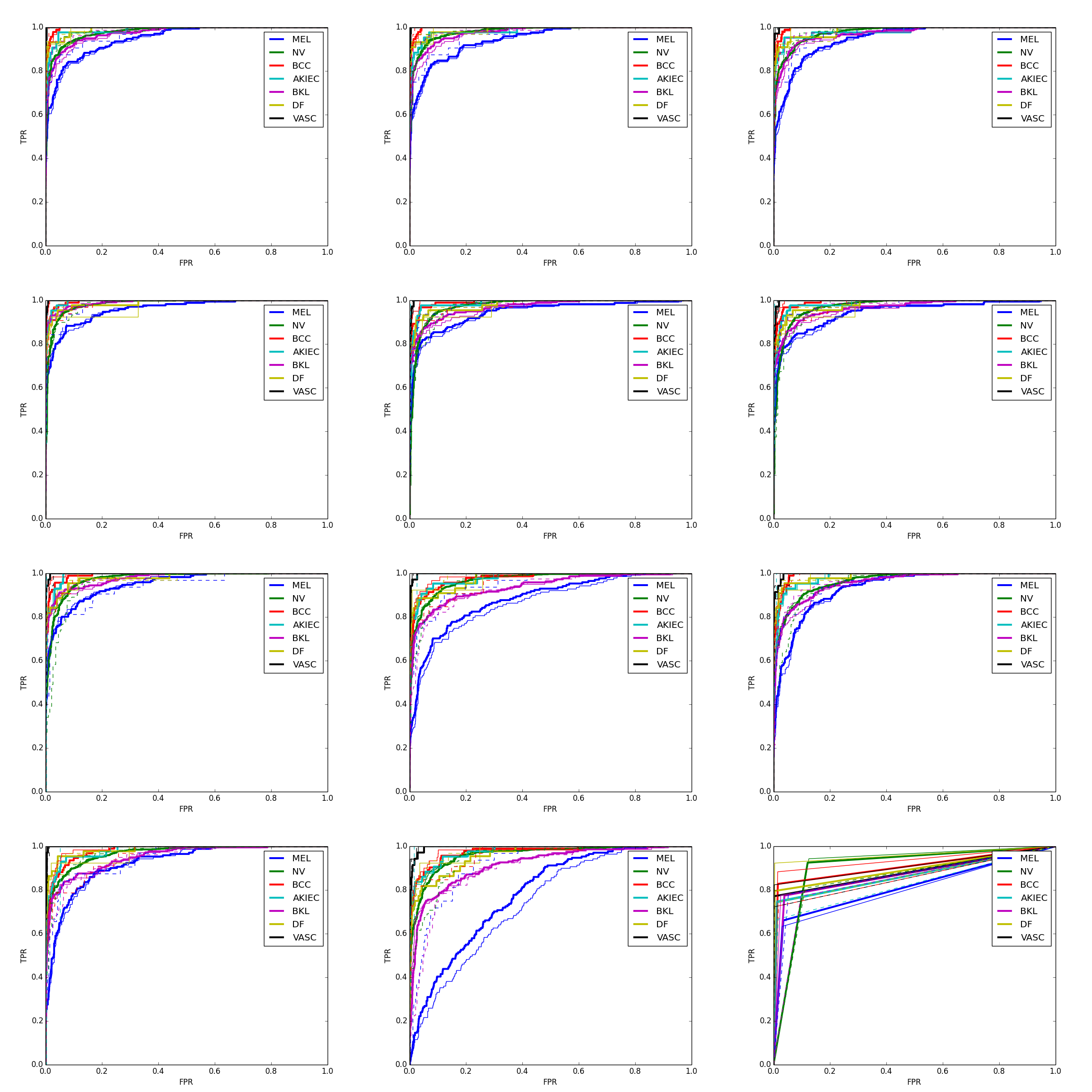}}
  \caption{ ROC plots of all 7 disease states for the top 12 performing submissions. Thick solid lines represents entire test dataset. Thin solid lines represents internal test split, and dotted thin lines represent external test splits. Submission rank in row order. }
\label{fig:roc_disease}
\end{figure}

\begin{figure}[h!]
  \centerline{\includegraphics[width=10cm]{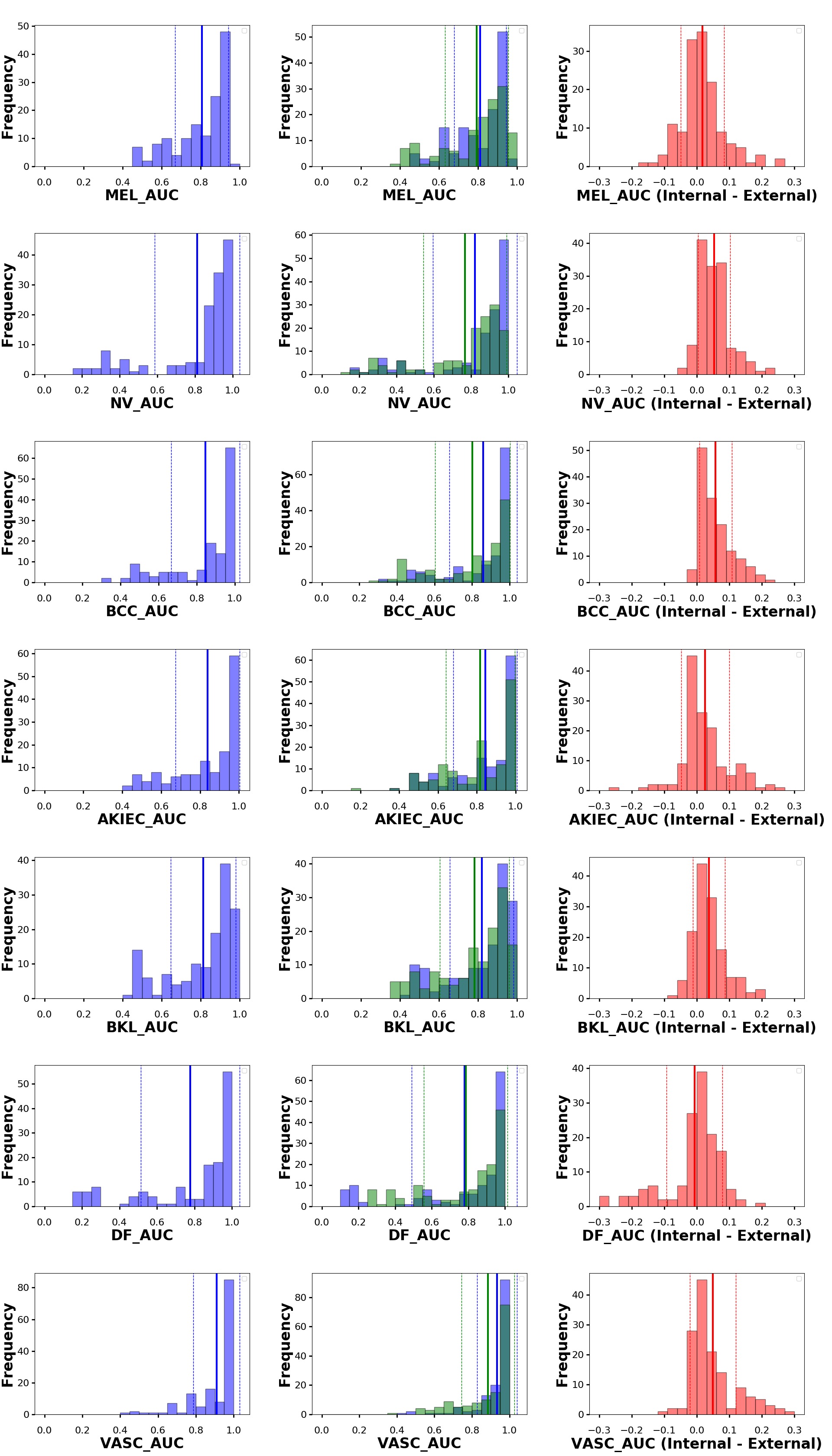}}
  \caption{ Histograms of submissions for Part 3: Lesion Classification, according to AUC for each disease category. Average values showed as solid vertical lines, and +/- standard deviation showed as dotted vertical lines. {\em Left Column:} Entire test dataset. {\em Center Column:} Internal (blue) and External (green) test partitions split. {\em Right Column:} Histogram of the subtraction of external test performance from internal. }
\label{fig:class_hist}
\end{figure}

\end{document}